\documentclass[12pt]{article}

\usepackage{amsmath,amsthm, amsfonts, amssymb, amsxtra, amsopn}
\usepackage{pgfplots}
\pgfplotsset{compat=1.13}
\usepackage{pgfplotstable}
\usepackage{graphicx,grffile}
\usepackage{multirow}
\usepackage{booktabs}
\usepackage{listings}
\usepackage{cmap}
\usepackage{colortbl}
\usepackage{adjustbox}
\usepackage{epsfig}
\usepackage{pifont}

\usepackage[tableposition=top,font=small,skip=5pt]{caption}
\usepackage{subcaption}
\usepackage{makecell} 

\newcommand{\cmark}{\ding{51}}%
\newcommand{\xmark}{\ding{55}}%

\PassOptionsToPackage{hyphens}{url}

\usepackage{hyperref}
\hypersetup{colorlinks=true,linkcolor=black,citecolor=black,urlcolor=blue,filecolor=black}
\hypersetup{pdfpagemode=UseNone,pdfstartview=}

\usepackage{enumitem}
\setlist[itemize]{noitemsep, topsep=0pt}

\advance\oddsidemargin by -0.35in
\advance\textwidth by 0.7in

\advance\topmargin by -0.3in
\advance\textheight by 0.6in

\long\def\symbolfootnotetext[#1]#2{\begingroup%
\def\thefootnote{\fnsymbol{footnote}}\footnotetext[#1]{#2}\endgroup}

\title{An Empirical Analysis of the Shift and Scale Parameters in BatchNorm}

\author{Yashna Peerthum\footnotemark[1]\ \ \ 
Mark Stamp\footnotemark[1]\,\,\footnotemark[2]}

\begin{document}

\symbolfootnotetext[1]{Department of Computer Science, San Jose State University}
\symbolfootnotetext[2]{mark.stamp$@$sjsu.edu}

\maketitle

\abstract
Batch Normalization (BatchNorm) is a technique that improves the training of deep neural networks, 
especially Convolutional Neural Networks (CNN).  
It has been empirically demonstrated that BatchNorm increases performance, stability, and accuracy,
although the reasons for such improvements are unclear. 
BatchNorm includes a normalization step as well as trainable shift and scale parameters. 
In this paper, we empirically examine the relative contribution to the success of BatchNorm
of the normalization step, as compared to the re-parameterization via shifting and scaling. 
To conduct our experiments, we implement two new optimizers in PyTorch, namely,
a version of BatchNorm that we refer to as AffineLayer, which includes the 
re-parameterization step without normalization,  
and a version with just the normalization step, that we call BatchNorm-minus. 
We compare the performance of our AffineLayer and BatchNorm-minus implementations
to standard BatchNorm, and we also compare these to the case where no batch normalization is used. 
We experiment with four ResNet architectures (ResNet18, ResNet34, ResNet50, and ResNet101) 
over a standard image dataset and multiple batch sizes. 
Among other findings, we provide empirical evidence that the success of BatchNorm 
may derive primarily from improved weight initialization.

\section{Introduction}\label{sect:intro}

In recent years, computational advances have resulted in the ability to train deeper networks, 
which has led to increases in classification accuracy on many tasks~\cite{deepnetworkproperties}. 
This has allowed for the use of deep learning for problems that had previously been considered extremely difficult. 
One such example is the use of deep Convolutional Neural Networks (CNN) for image classification. 
CNNs apply filters---also called kernels---to images to extract high level features such as edges. 
Modern CNNs can use many such layers, which can increase their classification accuracy. However, 
deeper networks come with a set of challenges~\cite{difffeedforward}. For example, in deep networks,
overfitting is often problematic, and convergence can become harder to achieve. In addition, deeper networks 
are more likely to result in so-called vanishing and exploding gradients. These gradient issues arise due 
to products of weights that are computed. In order to avoid these pitfalls, a number of approaches have been 
suggested with, arguably, the most notable being Batch Normalization (BatchNorm). 

In a CNN, BatchNorm can be viewed as a layer that we insert between convolutional layers. 
In effect, BatchNorm is a statistical regularization process, where we obtain a standardized output 
for each node in a layer. The mean~$\mu$ and standard deviation~$\sigma$ for 
a given neuron in a given layer
is determined over an entire batch. Then we normalize so that~$(\mu,\sigma)=(0,1)$, and we
compute the~$z$-score. This~$z$-score measures how far---in terms 
of standard deviation units---the input to the neuron is from the norm. The~$z$-score is then multiplied by 
a parameter~$\gamma$, added to another parameter~$\beta$, and the
result of this affine transformation is passed on to the next layer. 
The parameters~$\gamma$ and~$\beta$ are learned via training through backpropagation, 
along with the CNN weights. 

When BatchNorm was proposed in~2015, it was hailed as a breakthrough that unlocked the ability to 
develop networks with much greater complexity that did not experience the degradation in performance 
that had previously been observed~\cite{BatchNormics}. The use of BatchNorm yields an empirical increase in training accuracy, 
a reduction in the number of training steps, and it improves overfitting, in the sense of decreasing the number of 
dropout layers that are required. However, it is not clear why BatchNorm produces such improvements. 
The original authors of BatchNorm plausibly suggested that the reason for the success of the technique
was a reduction in the Internal Covariate Shift (ICS)~\cite{BatchNormics}. 

ICS is defined as ``the change in the distribution of network activations due to the change in 
network parameters during training''~\cite{BatchNormics}. In effect, each layer ``sees'' its input
as statistically differently, and the more layers, the greater this effect is likely to be.
In backpropagation, such differences 
can lead to inconsistent updates---when gradient descent is used to modify
the weights of a layer, there is an implicit assumption that the other layers have 
remained static~\cite{Goodfellow-et-al-2016}.

In its original formulation, a BatchNorm layer was placed before an activation function, 
which would serve to make the input to each layer statistically similar, thus
reducing ICS and thereby improving the representational power of the overall network. 
However, there is empirical evidence that placing the BatchNorm layer after the activation function achieves 
better results. This contradicts the original justification for BatchNorm, since the statistics of the inputs are not normalized. 
One attempt at analyzing BatchNorm injected noise into the data before the activation layers to skew the statistics 
of the layer---and thereby induce ICS---and found that there was no significant reduction in training 
accuracy as compared to standard BatchNorm~\cite{BatchNormnoics}. 
The authors of~\cite{BatchNormnoics} argue that what BatchNorm is really doing is re-parameterizing 
the model in a way that serves to smooth the loss surface and thereby accelerate convergence.
They further posit that ICS may not even be an issue that needs to be addressed 
when training a neural network. Another study claimed that the speed and stability of 
BatchNorm are independent effects~\cite{BatchNormspeedstabiliy}.

There are many alternatives to BatchNorm, with the normalization procedure taking place over 
differing attributes, e.g., normalization over layers, weights, and color channels each provide improvement,
as compared to a standard model with no normalization.
Many of these BatchNorm alternatives include shift and scale parameters,
which serve to re-parameterize the data~\cite{adabn,switchnorm,weightnorm,instancenorm,layernorm}. 

In this paper, we investigate the relationship between the re-parameterization
in BatchNorm, and the normalization itself. 
The shift and scale parameters provide two additional trainable parameters---and hence
two additional degrees of freedom---for each layer,
which could account for some of the improvements observed in BatchNorm. 
To isolate the role of these parameters, we have produced a package that we
call AffineLayer, which foregoes the normalization in BatchNorm
and only performs the shift and scale re-parametrization step.

We have also produced a package that we call BatchNorm-minus
that includes the normalization step of BatchNorm, but not the shift and scale
re-parameterization. By comparing our AffineLayer and BatchNorm-minus results to BatchNorm,
we hope to obtain insight into the relative contribution of the normalization step,
as compared to the re-parameterization step, with the ultimate goal of shedding further 
light on the mechanism by which BatchNorm improves the training of models.

The remainder of this paper is organized as follows.
In Section~\ref{chap:background}, we discuss relevant background topics, 
including the development of BatchNorm and previous attempts that have been made to 
understand why it works. We also discuss several alternatives to BatchNorm that have been 
previously developed. Section~\ref{chap:experiments_design} focuses on our
experimental design and provides implementation details.
In Section~\ref{chap:experiments_results}, 
we give our experimental results and discuss our findings. 
We conclude the paper in Section~\ref{chap:conclusion}, 
where we also consider directions for possible future work.

\section{Background}\label{chap:background}

This section deals with the details of BatchNorm, the principles behind it, 
some potential reasons for why it works, and alternatives to BatchNorm. 
We also discuss why ResNet was chosen for
this research and the contributions of this research paper.

\subsection{How BatchNorm Works}

BatchNorm is an optimization and re-parametrization procedure performed per mini-batch. It consists of a normalization 
step that is based on the statistics of each mini-batch, along with per-layer shift and scale parameters. 
During the validation phase, the shift and scale parameters from the training phase are used to evaluate inputs.

\subsubsection{Mathematics of BatchNorm}

BatchNorm operates on every individual feature per mini-batch. For each feature~$x_j$, we calculate
the mean and variance as
$$
 \mu_j = \frac{1}{m} \sum_{i = 1}^{m} x_{ij} \mbox{\ \ and\ \ } {\sigma_j}^2 = \frac{1}{m} \sum_{i = 1}^{m} {(x_{ij} - \mu_j)}^2
$$
respectively. We then uses these values to determine the~$z$-score
$$
\widehat{x}_j = \frac{x_{ij} - \mu_j}{\sqrt{{\sigma_j}^2 + \varepsilon}}
$$
where~$\varepsilon$ is a constant for stabilizing the output
Finally, we calculate the output
$$
y_j = \gamma_{\ell}\kern 1pt\widehat{x}_j + \beta_{\ell}
$$
where~$\gamma_{\ell}$ and~$\beta_{\ell}$ are per-layer parameters that are updated through 
backpropagation, alongside the weights. 

According to the original BatchNorm paper~\cite{BatchNormics},
the purpose of the scale and shift parameters~$\gamma$ and~$\beta$ is to restore information that may be lost through 
the normalization process of zeroing the mean. This scale and shift serves to re-parametrize the activations 
in a way that allows for the same family of functions to be expressed, but with
trainable parameter that may make it easier to learn via gradient descent~\cite{Goodfellow-et-al-2016}.

\subsubsection{Internal Covariate Shift}

Internal Covariate Shift (ICS) is a known issue for deep neural nets. 
The more layers the neural network has, the more the statistics of each layer change 
as a result of the preceding layers and, more specifically, the distribution of the activation 
functions change as more updates are performed. In backpropagation, 
when parameters are updated, all layers are updated simultaneously,
under the implicit assumption that the statistics of previous layers are static~\cite{Goodfellow-et-al-2016}.
The ``covariate'' refers to the way that the inputs to the neural network vary relative to each other, 
while the ``shift'' refers to the change in the distributions of the outputs across different layers. 
The effect of ICS on statistical predictions predates neural networks~\cite{icsoriginal}.
In the context of neural networking, ICS has also been discussed with respect to
distribution changes between the training and validation domains~\cite{icsdomain}.

\subsubsection{What is BatchNorm Trying to Achieve?}

There is no rigorous proof as to why BatchNorm improves the performance of deep neural networks,
nor is there a consensus on why the technique is so successful. 
BatchNorm has been shown empirically to regulate overfitting and reduce or eliminate the need for dropout layers. 

A large learning rate can cause erratic training 
behavior, but a learning rate that is too small can cause 
a model to fail to converge or to require more training epochs.
BatchNorm allows for a smaller learning rate with no loss in performance,
and it seems to perform better with a batch size of~30 or more.
The creators of BatchNorm posited that it could be reducing the ICS and hence it works by 
reducing the ``randomness'' that occurs in each batch of data
as a consequence of simultaneous updates over many layers~\cite{BatchNormics}. 
 
Where to place the BatchNorm layers within a CNN has been a topic of much debate. 
The original paper places BatchNorm layers before the activation function, which is supposed to 
improve ICS and thereby reduce vanishing or exploding gradient issues during training. 
However, in practice it has been observed that BatchNorm performs 
as well---or even better---when it is applied to the output of the activation function.
This would seem to contradict the stated purpose of the parameters~$\beta$ and~$\gamma$, which 
are supposed to control the statistics of the layer, since that ability is reduced when the output is passed 
through the activation function prior to BatchNorm. 

There are published research papers, including~\cite{icsbound,BatchNormics,icsdomain}, 
that argue that BatchNorm works 
because it reduces ICS and acts like a normalizer. However, there are other papers,
such as~\cite{decorrbn,BatchNormnoics},
that provide evidence that ICS is not even a real
issue for neural networks, and that BatchNorm is just smoothing the gradient landscape and 
thereby accelerating convergence. Part of the evidence for this latter perspective
is that we can place the BatchNorm layer anywhere and obtain improvements, as compared to no such normalization. 
Others make the argument that BatchNorm smoothes out the loss surface, which makes it easier to reach 
a global maximum~\cite{BatchNormnoics}.

In this paper, we conduct experiments that attempt to separate out the relative effect of the
two components of BatchNorm, namely, the normalization step and the re-parameterization 
via the trainable parameters~$\gamma$ and~$\beta$.
By considering these components separately, as detailed in Section~\ref{sect:over} below,
we hope to gain insight into how and why BatchNorm is so effective.
As far as the authors are aware, this is a novel approach to analyzing BatchNorm.

\subsection{Datasets}

\begin{figure}[!htb]
    \centering
    \includegraphics[width=0.625\textwidth]{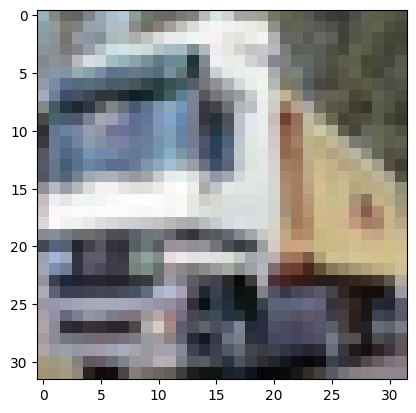}
    \caption{Example of CIFAR10 image of truck}\label{fig:cifar10-truck}
\end{figure}

The Canadian Institute for Advanced Research 10-class dataset, commonly referred to as CIFAR10~\cite{cifar}, 
is comprised of~60,000 images, each of which is a~$32\times 32$ pixel image. 
Alongside ImageNet, it is one of the most widely used datasets for machine learning and deep learning
research. The CIFAR10 dataset consists of~50,000 training images (5000 from each of the~10 classes), 
and~10,000 validation images (1000 from each class). 
The~10 mutually exclusive classes are airplane, automobile, bird, cat, deer, dog, frog, horse, ship, and truck.
Also, each image only contains the specified object. This dataset is available as part of the
PyTorch~\cite{pytorch} package. 
An example of a CIFAR10 image is given in Figure~\ref{fig:cifar10-truck}.

In this paper, we use the CIFAR10 
dataset for our experiments. 
CIFAR10 has been considered in several other research papers that attempt to evaluate 
BatchNorm.

\subsection{ResNet}

We use Residual Networks (ResNet) for our experiments, in part because 
ResNet was developed with BatchNorm built into it. 
Papers evaluating BatchNorm or its variants typically choose ResNet for this reason, 
as well as it being one of the current best classifiers~\cite{resnetperformance}.

ResNet was developed in~2015 and it can be viewed as a type of Convolutional Neural Network (CNN).
As alluded to above, ResNet is considered the current state of the art in computer vision~\cite{resnet},
as it consistently outperforms its competitors in competitions involving image tasks. 
A ResNet architectures uses a series of repeating blocks made from convolution layers that are 
designed to model the ``residual function'' instead of the output. Here, the residual function is defined 
as the difference between the input and the output of a residual block. If the input is~$x$ and the output 
is~$F(x)$ then the residual being modeled is~$F(x) - x$. The motivation behind ResNet was the observation
that deeper networks sometimes performed worse than shallower networks, which is counterintuitive, as
``excess'' layers should simply model the identity function. By modeling the residual,
ResNet blocks model the identity function as~0, which may be easier to learn
via backpropagation.

Each residual block in a ResNet has two pathways, namely, the residual mapping or an identity mapping,
where the identity mapping implies that the block is, in effective, being skipped over,
thus improving gradient flow through the remainder of the network.
This makes deeper networks more feasible---ResNets can be trained with vastly more 
layers as compared to traditional neural networks. 
Residual blocks also serve to decrease overfitting. 
In practice, ResNet appears to act as a collection of shallow neural nets---in effect, an ensemble that is trained
simultaneously~\cite{resnetperformance}. 

ResNet uses Rectified Linear Unit (ReLu) activations, which have been 
empirically shown to decrease the incidence of vanishing gradients. In addition, ResNet makes use of BatchNorm 
before each activation function. Each specific ResNet architecture is 
specified as ``ResNet\kern 1pt$n$'', where~$n$ denotes the 
number of convolution layers in the ResNet. Popular ResNet architectures
include ResNet18, ResNet34, ResNet50, ResNet101, ResNet110, ResNet152, ResNet164, and ResNet1202.
Note that, in general, the residual block structure in different ResNet architectures differ.

As mentioned above, 
ResNet architectures utilize BatchNorm 
within residual blocks,
and the contribution of BatchNorm to the success of these 
architectures is well documented.
These factors makes ResNet an ideal candidate for our experiments, 
where we attempt to understand the relative contributions of the components
of BatchNorm (i.e., normalization and re-parameterization).

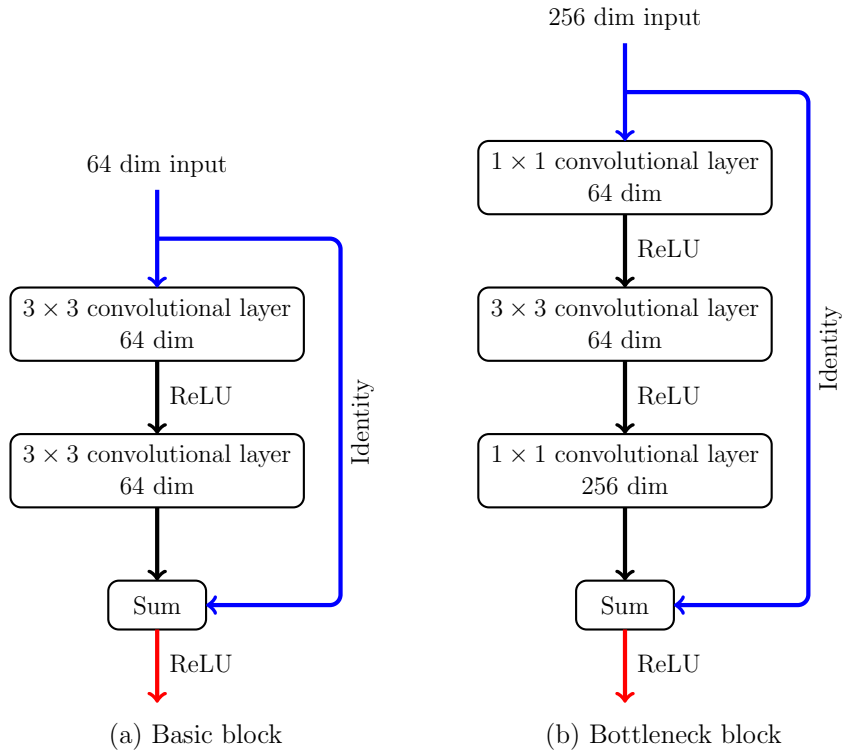
\begin{figure}[!htb]
    \centering
    \begin{tabular}{cccc}
        \begin{tikzpicture}[scale=0.65,every node/.style={scale=0.8}]

    \draw[black,thick,rounded corners] (0.5,0.0) rectangle (2.5,1.0);
    \node at (1.5,0.5) {Sum};

    \draw[black,thick,rounded corners] (-1.5,2.5) rectangle (4.5,4.0);
    \node at (1.5,3.55) {$3\times 3$ convolutional layer};
    \node at (1.5,2.95) {64 dim};

    \draw[black,thick,rounded corners] (-1.5,5.5) rectangle (4.5,7.0);
    \node at (1.5,6.55) {$3\times 3$ convolutional layer};
    \node at (1.5,5.95) {64 dim};

    \node at (1.5,9.5) {64 dim input};

    \draw[red,ultra thick,->] (1.5,0) -- (1.5,-1.5);
    \draw[black,ultra thick,->] (1.5,2.5) -- (1.5,1);
    \draw[black,ultra thick,->] (1.5,5.5) -- (1.5,4);
    \draw[blue,ultra thick,->] (1.5,9) -- (1.5,7);

    \node at (2.4,-0.7) {ReLU};
    \node at (2.4,4.8) {ReLU};

    \draw[blue,ultra thick,rounded corners,->] (1.5,8.0) -- (5.25,8.0) -- (5.25,0.5) -- (2.5,0.5);
    \node[rotate=90] at (5.7,4.25) {Identity};

     
    \end{tikzpicture}
    & & &
        \begin{tikzpicture}[scale=0.65,every node/.style={scale=0.8}]

    \draw[black,thick,rounded corners] (0.5,0.0) rectangle (2.5,1.0);
    \node at (1.5,0.5) {Sum};

    \draw[black,thick,rounded corners] (-1.5,2.5) rectangle (4.5,4.0);
    \node at (1.5,3.55) {$1\times 1$ convolutional layer};
    \node at (1.5,2.95) {256 dim};

    \draw[black,thick,rounded corners] (-1.5,5.5) rectangle (4.5,7.0);
    \node at (1.5,6.55) {$3\times 3$ convolutional layer};
    \node at (1.5,5.95) {64 dim};

    \draw[black,thick,rounded corners] (-1.5,8.5) rectangle (4.5,10.0);
    \node at (1.5,9.55) {$1\times 1$ convolutional layer};
    \node at (1.5,8.95) {64 dim};

    \node at (1.5,12.5) {256 dim input};

    \draw[red,ultra thick,->] (1.5,0) -- (1.5,-1.5);
    \draw[black,ultra thick,->] (1.5,2.5) -- (1.5,1);
    \draw[black,ultra thick,->] (1.5,5.5) -- (1.5,4);
    \draw[black,ultra thick,->] (1.5,8.5) -- (1.5,7);
    \draw[blue,ultra thick,->] (1.5,12) -- (1.5,10);

    \node at (2.4,-0.7) {ReLU};
    \node at (2.4,4.8) {ReLU};
    \node at (2.4,7.8) {ReLU};

    \draw[blue,ultra thick,rounded corners,->] (1.5,11.0) -- (5.25,11.0) -- (5.25,0.5) -- (2.5,0.5);
    \node[rotate=90] at (5.7,5.75) {Identity};

     
    \end{tikzpicture}
    \\
    \adjustbox{scale=0.85}{(a) Basic block} 
    & & &
    \adjustbox{scale=0.85}{(b) Bottleneck block} 
    \end{tabular}
    \caption{Residual blocks in selected ResNet architectures}\label{fig:ResNet}
\end{figure}

In this research, we consider ResNet18, ResNet34, ResNet50, and ResNet101. 
All of these achieve reasonably high accuracies on the CIFAR10 dataset. 
These architectures use different residual blocks: ResNet18 and ResNet34 
use smaller two-convolution layer blocks called \textit{basic blocks},
while ResNet50 and ResNet101 use three-convolution layer blocks called 
\textit{bottleneck blocks}. 
The bottleneck blocks in ResNet50 and ResNet101 reduce the dimensionality 
of the~256 dimensional input before 
applying~$3\times 3$ convolution filters, 
the result of which is then projected back into~256 dimensions. 
Basic blocks and bottleneck blocks are illustrated in 
Figures~\ref{fig:ResNet}(a) and~(b), respectively.

Since these four architectures perform comparably well on the CIFAR datasets, they will help
us to determine whether our empirical results are dependent on the depth of the ResNet.
Also, since there are two different block structures among the four architectures under
consideration, we can observe the effect of the block type, relative to the components of BatchNorm.

\subsection{Related Work}

In this section, we consider relevant related research that has attempted to demystify BatchNorm, 
summarizing the methods used and the results that were obtained. We also 
consider a few BatchNorm alternatives and discuss how they achieve their results.

\subsubsection{Refuting ICS}

In the~2018 paper~\cite{BatchNormnoics}, evidence is provided that the success of BatchNorm is not due to ICS at all,
and that it might not even be reducing ICS. To support this hypothesis, based on a VGG classifier and CIFAR10, 
they injected noise into the activations following each BatchNorm layer so that any mitigation of ICS that BatchNorm
might have provided was no longer valid. This noise was randomized at each time step so that none of the distributions 
were identical. These experiments showed that there was no drop in performance between the noise-injected network 
and the one with standard BatchNorm. In addition, both networks significantly outperformed a VGG neural net 
that did not use BatchNorm. This contradicts the original BatchNorm paper;
if ICS reduction was the main benefit of BatchNorm, then adding ICS to 
a network would degrade its performance. In contrast, these experiments 
show that BatchNorm improves the performance of VGG, even when ICS is increasing. 

In the paper~\cite{BatchNormnoics},
the authors also proposed a more precise definition for ICS in the context of neural networking. 
ICS should be reflected in how much a neural networking layer needs to adapt to changes in its inputs.
Therefore, they quantify ICS as the difference between~$G$ and~$G'$, where~$G$ consists of
the gradient parameters before updates and~$G'$ consists of these same set of parameters after the updates. 
They measured this difference in terms of the~$L_2$ norm and cosine similarity. 

If BatchNorm was indeed reducing ICS, as defined by the metric in the previous paragraph, 
then the use of BatchNorm layers would decrease the correlation between~$G$ and~$G'$ since 
BatchNorm would cause less cross-layer dependency. However, the authors of~\cite{BatchNormnoics} 
obtained the surprising result that a network with BatchNorm was increasing the correlation between 
these variables. The authors believe that this occurs because BatchNorm effectively re-parameterizes 
the loss function, and hence its impact is likely due to an improvement in this surface. 
That is, BatchNorm could be smoothing the loss surface, making it easier for gradient descent 
to reach the desired outcome. To verify this, they considered the ``Lipschitzness'' of the loss function 
with and without BatchNorm. 

A function~$f$ is~$K$-Lipschitz provided that
$$
\| f(x_1) - f(x_2)\|  \leq K \| x_1 - x_2 \|
$$
for all choices of~$x_1$ and~$x_2$, where~$K$ is real-valued, with~$K>0$. 
Lipschitz continuity means that the function is strongly uniform continuous and~$K$ provides a 
limit on how rapidly it can change. A larger~$K$ implies a smoother loss function, which is a good thing
with respect to convergence via gradient descent. 

The authors of~\cite{BatchNormnoics} discovered that 
BatchNorm did not just improve the Lipschitzness of the loss function but also the Lipschitzness of its gradients, 
which implies increased convexity. This makes a strong case for the underlying mechanics of BatchNorm, 
but it does not settle which aspects of BatchNorm lead to the improvements. It could simply be the case
that the addition of more trainable parameters (i.e., shift and scale) helps, since it is theorized 
that having more parameters might explain some of the performance disparity between 
shallow and deep networks~\cite{deepnetworkproperties}.

\subsubsection{Further Experiments on the ICS}

In the~2020 paper~\cite{icsbound}, the authors claim to improve BatchNorm by using an alternative metric for ICS, 
and they determine upper and lower bounds for the ICS. They use the so-called earth mover's distance, 
a measure that quantifies the distance between two probability distributions. In this case,
the distributions being compared are based on the statistics of the gradient values before and after updates. 
The paper claims to obtain an improvement over BatchNorm, albeit a small one.
Interestingly, their normalization step involves an additional parameter~$\alpha$,
which is trained alongside the~$\beta$ and~$\gamma$ parameters. 
They tested their algorithm on various ResNets~\cite{earthmovers}.

\subsubsection{Regularization and BatchNorm}

In the~2018 paper~\cite{BatchNormregul}, population statistics were used instead of batch statistics, 
and a regularization term was included for the~$\gamma$ parameter. In these experiments, it was noted that 
for batches of size larger than~32, the population statistics function as well as the batch statistics. 
It was also noted that BatchNorm introduces Gaussian noise into the mean and variance parameters. 
The algorithm in~\cite{BatchNormregul} produces accuracies that are comparable to BatchNorm, 
but unlike the case with BatchNorm, they found that introducing dropout layers improved 
performance further. They also state that BatchNorm 
has very similar effects to~$L^p$ norm regularization where the~$p$-norm is defined as
$$
  \|x_p\| = \Bigl(\sum_{i=1}^{n} |x_i|^p\Bigr)^{1/p} \!.
$$

If~$L^p$ norm regularization was sufficient, there would be no need for an optimization such as BatchNorm,
since regularization is less computationally intensive. However, this regularization claim is not consistent with
other empirical studies, such as that presented in the~2018 paper~\cite{weightnormreg}.

\subsubsection{Weight Normalization and BatchNorm}

While BatchNorm deals with the input to activation functions in a layer, it would be reasonable to
attempt to normalize the weights of a layer directly~\cite{weightnorm}. This WeightNorm
approach works per batch, similar to BatchNorm but claims to be less noisy and more computationally 
efficient, especially for shallower networks. In WeightNorm, the weight vector is re-parametrized as
$$
w = \frac{g}{\|v\|} v
$$
where $g$ is a trainable parameter and~$\|v\|$ is the Euclidean norm of~$v$.
The authors of~\cite{weightnorm} combine this with a form of BatchNorm, meant only to center the gradients.
This is referred to as ``mean-only BatchNorm'', where only the mean of the neuron inputs is calculated. 
The authors of WeightNorm emphasize that one advantage of their normalization 
is that it decouples the direction of the weight vector from its magnitude, 
and this has led to speculation that the performance of BatchNorm is also
due to this property~\cite{decoupling}.

\subsubsection{Decoupling the Length and Direction of the Weights}

In~\cite{decoupling}, it is shown that the transformations that BatchNorm imposes results in the magnitude of the weight vector 
being independent of the direction of the vector. The authors hypothesize that this allows BatchNorm to use 
properties of the optimization landscape in a way that other regularization methods cannot. Using this property in their 
optimization step, they were able to achieve a linear convergence on a non-convex problem. 

\subsubsection{Residual Learning without Normalization}

Another way in which BatchNorm could be improving deep neural nets is by making the weight initializations 
more consistent at the start of each epoch. The weight initialization problem had been discussed before 
the inception of BatchNorm~\cite{difffeedforward}. In~2019, a paper based on research done at Facebook 
developed a method called Fixup Initialization~\cite{fixup}. This method is another attempt to solve the 
exploding and vanishing gradient issue, which is related to the fact that the deeper the neural net, 
the larger the variance of its output will tend to be. Fixup introduces a rescaling of the standard weight initializations 
and it also includes trainable shift and scale parameter similar to BatchNorm.
The authors claim that using their Fixup, they can obtain results that are superior to BatchNorm on the 
CIFAR10 dataset, based on a ResNet architecture. Given that they also use shift and scale parameters, 
this does not have clear implications for the effective mechanism behind BatchNorm.

\subsubsection{Decorrelated Batch Normalization}

One of the motivations behind the development of BatchNorm was the idea that inputs to the activations 
should be whitened, which requires scaling, standardizing, and decorrelating. BatchNorm however only implements the first two, 
because it is computationally intensive to decorrelate---this would require computing the inverse square root 
of the covariance matrix during back propagation~\cite{whitening}. The~2018 paper~\cite{decorrbn} implements 
what the authors call ``Decorrelated BatchNorm'' through a process called Zero Phase Component Analysis. 
This process involves scaling along eigenvectors, and is similar to Principal Component Analysis,
except that it but does not rotate the coordinate axes. The authors use the transformation
$$
  \widehat{x}_i = \raisebox{-1.0pt}{$\Sigma^{1/2}$}\bigl(x_i - \mu\bigr)
$$
where~$\mu$ is the mean of the mini-batch and~$\Sigma$ is the covariance matrix of the mini batch.

Testing on CIFAR10, and the more challenging CIFAR100 dataset, 
using ResNet the authors of~\cite{decorrbn} 
note that the whitening process creates an improvement in performance over vanilla BatchNorm. 
They also recommend including shift and scale parameters, since these also improved performance. 
However, the computational cost of whitening is non-trivial.

\subsubsection{Adaptive Batch Normalization}

Adaptive Batch Normalization (AdaBN) improves on BatchNorm in 
transfer learning applications~\cite{adabn}. One of the issues with BatchNorm is that there
is a disconnect between source and target domains, 
in the sense that the statistics used for each differ.
Here, the source is the data the weights are 
derived from, while the target is the new data that we are classifying. 
AdaBN uses the BatchNorm statistics and combines them with weight statistics,
with the rationale being that BatchNorm statistics provide information about the source, while the weight statistics 
provide information about the target. For neuron~$j$ and for an image~$m$ in the dataset, they calculate
$$
y_j(m) = \gamma_j \frac{x_j(m) - \mu_j^t}{\alpha_j^t} + \beta_j
$$
where~$\mu_j^t$ and~$\alpha_j^t$ are, respectively, 
the mean and variance of the outputs of the neuron in the target domain.

\subsubsection{AutoDIAL: Automatic DomaIn Alignment Layers}

Another transfer learning algorithm is AutoDIAL~\cite{autodial}, which attempts to maximize classification accuracy 
by aligning the source and target domains. They do so by looking at statistics from both domains in advance 
and designing a parameter~$\theta$ that represents the shared weights. They still use BatchNorm layers 
to bring the two domain together but they do so via a parameter~$\alpha$ that quantifies the degree of 
mixing of both sets of statistics. If~$\alpha=1$, then the domains are not aligned while~$\alpha=0.5$ 
indicates that they are partially aligned. 

\subsubsection{Layer Normalization}

LayerNorm functions within a mini-batch, where it is trying to normalize the inputs with respect to 
the other features in the same layer of the neural network~\cite{layernorm}. This approach uses the same 
statistics as BatchNorm but whereas BatchNorm is based on the same feature, 
LayerNorm is computed across different features. This works best when the features are similar 
to each other in scale. LayerNorm has been used successfully in Recurrent Neural Networks (RNN) 
and transformers-based machine learning models. Since it functions per layer, unlike BatchNorm, there are no dependencies 
between layers and hence LayerNorm would not be expected to result in any decrease of ICS within a network.

\subsubsection{Instance Normalization}

InstanceNorm is a variation of LayerNorm that works across RGB channels instead of 
features~\cite{layernorm}. This is an attempt at maximizing contrast within images and it
has been applied with success to GANs. 

\subsubsection{Group Normalization}

GroupNorm was created to allow for smaller batch sizes, as compared to 
standard BatchNorm~\cite{groupnorm}. 
For high-resolution images, smaller batches of size one or two are preferred,
whereas BatchNorm requires larger batch sizes to perform well. GroupNorm, does not normalize 
in batches, but instead normalizes along the feature dimension by considering groups of features. 
GroupNorm has been shown to work well
for batches of size two, and it may enhance object segmentation and detection. 

\subsubsection{SwitchBlade Normalization}

SwitchBlade Normalization (SN) combines three approaches that we have 
discussed above~\cite{switchnorm}. Specifically, SN combines 
InstanceNorm (to normalize across each feature), 
LayerNorm (to normalize across each layer), 
and BatchNorm (to normalize across each batch).
The algorithm learns which of the three types of normalizations works best with the data and 
can ``switch'' between any combination of the three that achieves the best result. 
The authors of SN note that of the three normalizations, BatchNorm is assigned
the highest weight during image classification tasks.

\section{Experimental Design}\label{chap:experiments_design}

In this section, we present our experimental process from a high-level perspective. 
Specifically, we discuss the design of our
BatchNorm variants, our PyTorch implementations,
and the hyperparameters selected for the experiments
presented in Section~\ref{chap:experiments_results}.

\subsection{Architecture Selection}

Above, we explained that ResNet was chosen for our experiments because it is the current best image classifier 
and it is heavily dependent on BatchNorm. ResNet also comes in multiple variants, enabling
us to easily experiment with different depths and different residual block structures. 
Based on preliminary tests, we chose to focus on ResNet18, ResNet34, ResNet50, and ResNet101, 
since these models are fast to train, and they are sufficient to illustrate 
the key points of our research. Recall that ResNet18 and ResNet34 use basic residual blocks,
while ResNet50 and ResNet101 use bottleneck blocks, as
illustrated in Figure~\ref{fig:ResNet}, above. Hence, ResNet34 can be viewed as a deeper
version of ResNet18, and the same can be said of the pair ResNet101 and ResNet50.
However, ResNet50 is not just a deeper version of ResNet18, for example. 

\subsection{BatchNorm Variants}\label{sect:over}

As mentioned above, we have implemented two BatchNorm variants 
that are designed to help us determine the relative contributions of the normalization step, 
as compared to the re-parameterization step. The first of these, which we refer to as AffineLayer,
includes only the affine transformation part of BatchNorm. That is,
AffineLayer does not normalize the output, but does include trainable shift and scale parameters 
($\gamma$ and $\beta$, respectively). These parameters offer two additional degrees of 
freedom for each neuron, which may allowing for more expressive 
models. 
We also develop and analyze a variant that includes the normalization step of BatchNorm, but not
the re-parameterization, which we refer to as BatchNorm-minus. 
We compare these techniques to standard BatchNorm and to the case where
no normalization is use, which we denote as ``none'' in our tables and graphs.
Table~\ref{tab:permutations} summarizes the four variations that we
test on the ResNet18, ResNet34, ResNet50, and ResNet101 networks.

\begin{table}[!htb]
\caption{Normalizations tested}\label{tab:permutations}
\centering
\adjustbox{scale=0.85}{
\begin{tabular}{c|cc}\midrule\midrule
\multirow{2}{*}{Normalization} & Re-parameterize & \multirow{2}{*}{Re-normalize} \\
 & ($\beta$ and~$\gamma$) & \\ \midrule
BatchNorm & \cmark & \cmark\\
AffineLayer & \cmark & \xmark \\
BatchNorm-minus & \xmark & \cmark \\
None & \xmark & \xmark \\ \midrule\midrule
\end{tabular}
}
\end{table}

Tables~\ref{tab:hyperparameters18}
and~\ref{tab:hyperparameters50} summarize
the hyperparameters tested (via grid search)
for the ResNet architectures under consideration. 
In both of these tables, boldface is used to indicate
the hyperparameter value that yields the best result.
Note that in every case, the Adam optimizer is best, as
is a learning rate of~$0.001$, while the best choice
for batch size varies considerably. 
Therefore, for all experiments in 
Section~\ref{chap:experiments_results}, we use the
Adam optimizer and a learning rate of~$0.001$, and all models
are trained for~15 epochs. 
To reduce the number of potential confounding variables,
we only consider batch sizes of~20 and~50.
Thus, our experiments will not necessarily yield the 
best possible accuracies, but that is not our purpose.
Instead, our goal to highlight differences 
between the ResNet models, relative to the four 
normalization schemes under consideration.

\begin{table}[!htb]
\caption{Hyperparameters for ResNet18 and ResNet34}\label{tab:hyperparameters18}
\centering
\adjustbox{scale=0.78}{
\begin{tabular}{c|ccc}\midrule\midrule
Normalization & Hyperparameter & Values tested & Best validation accuracy\\ \midrule
\multirow{3}{*}{BatchNorm} & Learning rate & 0.01,0.005,\textbf{0.001} & \multirow{3}{*}{0.7665}\\ 
& Optimizer & \textbf{Adam}, SGD \\ 
& Batch size & $20,30,\ldots,100$ (\textbf{30}) \\ \midrule
\multirow{3}{*}{AffineLayer} & Learning rate & 0.01,0.005,\textbf{0.001} & \multirow{3}{*}{0.6904}\\ 
& Optimizer & \textbf{Adam}, SGD \\ 
& Batch size & $20,30,\ldots,100$ (\textbf{80})\\ \midrule
\multirow{3}{*}{BatchNorm-minus} & Learning rate & 0.01,0.005,\textbf{0.001} & \multirow{3}{*}{0.7730}\\ 
& Optimizer & \textbf{Adam}, SGD \\ 
& Batch size & $20,30,\ldots,100$ (\textbf{20})\\ \midrule
\multirow{3}{*}{None} & Learning rate & 0.01,0.005,\textbf{0.001} & \multirow{3}{*}{0.6877}\\ 
& Optimizer & \textbf{Adam}, SGD \\ 
& Batch size & $20,30,\ldots,100$ (\textbf{80})\\ \midrule\midrule
\end{tabular}
}
\end{table}

\begin{table}[!htb]
\caption{Hyperparameters for ResNet50 and ResNet101}\label{tab:hyperparameters50}
\centering
\adjustbox{scale=0.78}{
\begin{tabular}{c|ccc}\midrule\midrule
Normalization & Hyperparameter & Values tested & Best validation accuracy\\ \midrule
\multirow{3}{*}{BatchNorm} & Learning rate & 0.01,0.005,\textbf{0.001} & \multirow{3}{*}{0.7469}\\ 
& Optimizer & \textbf{Adam}, SGD \\ 
& Batch size & $20,30,\ldots,100$ (\textbf{70})\\ \midrule
\multirow{3}{*}{AffineLayer} & Learning rate & 0.01,0.005,\textbf{0.001} & \multirow{3}{*}{0.6986}\\ 
& Optimizer & \textbf{Adam}, SGD \\ 
& Batch size & $20,30,\ldots,100$ (\textbf{80})\\ \midrule
\multirow{3}{*}{BatchNorm-minus} & Learning rate & 0.01,0.005,\textbf{0.001} & \multirow{3}{*}{0.6540}\\
& Optimizer & \textbf{Adam}, SGD \\ 
& Batch size & $20,30,\ldots,100$ (\textbf{100})\\ \midrule
\multirow{3}{*}{None} & Learning rate & 0.01,0.005,\textbf{0.001} & \multirow{3}{*}{0.6939}\\ 
& Optimizer & \textbf{Adam}, SGD\\ 
& Batch size & $20,30,\ldots,100$ (\textbf{70})\\ \midrule\midrule
\end{tabular}
}
\end{table}


 \subsection{Implementation}

We conduct our experiments using PyTorch, 
an open source machine learning library, which is itself based 
on the Torch package. For our purposes, the main benefit of PyTorch comes from its use of tensors, 
which are auto-differentiable numerical arrays that allow GPU parallelized 
Basic Linear Algebra Subprograms (BLAS) operations~\cite{pytorch}. 
This enables us to implement our AffineLayer as a tensor layer in the form of a custom 
PyTorch \texttt{nn.module}, which is one of the base classes. 
The two parameters, $\beta$ and~$\gamma$, are tensors and can be multiplied and added to the input layer. 
Defining~$\beta$ and~$\gamma$ as \texttt{nn.Parameter}, which is a type of array, makes them part 
of the computational graph which enables them to be trained via backpropagation. 
PyTorch is also convenient for our purposes because it offers a 
ResNet builder function that lets the user select which optimization 
layer to pass to the builder as an argument. 

To implement BatchNorm-minus, we modified the existing BatchNorm layer
in PyTorch to make the shift and scale no longer trainable, 
which leaves them at their initial values of~$(\beta,\gamma)=(0,1)$. 
To train our models without any type of normalization
(which is denoted as ``none'' in our tables and graphs),
we simply use an identity layer in place of BatchNorm.

Finally, all of our experiments have been run on an RTX3080Ti GPU and 
take on average two minutes per epoch to complete.
Since our hardware enables fast training, we are able to conduct 
a large number of experiments. 

\section{Experimental Results}\label{chap:experiments_results}

In this section, we compare the four different normalizations discussed above, 
namely, BatchNorm, AffineLayer, BatchNorm-minus, and ``none'' (i.e., no normalization).
First, we give results for the ResNet18 and ResNet50
architectures, then we consider ResNet34 and ResNet101. 
We also provide an in-depth analysis of gradient and weight statistics for our models,
and we conclude this section with a discussion of our experimental results.

\subsection{ResNet18 and ResNet50 Experiments}

As with all of our experiments, for the ResNet18 and ResNet50 models, we use the CIFAR10 dataset. 
These two architectures are the shallowest of their respective types, with ResNet18 using basic residual blocks 
and ResNet50 using bottleneck residuals blocks. Our experiments with these models provide a point of comparison 
between the two types of blocks and also enable us to see how depth affects the results. 
Following our experiments with these models, we further experiment
with the ResNet34 and ResNet101 architectures.

\subsubsection{ResNet18 Results}

Table~\ref{tab:182050} 
compares the four normalization---in terms of validation accuracy---based on the average of four 
different runs of each. The differences between batch sizes of~20 and~50 are marginal.

\begin{table}[!htb]
\caption{ResNet18 experimental results}\label{tab:182050}
\centering
\adjustbox{scale=0.85}{
\begin{tabular}{c|cc}\midrule\midrule
\multirow{2}{*}{Normalization} & \multicolumn{2}{c}{Validation accuracy}\\
 & Batch size~20 & Batch size~50\\ \midrule
BatchNorm & 0.7665 & 0.7569\\
AffineLayer & 0.6794 & 0.6904\\
BatchNorm-minus & 0.7730 & 0.7644\\
None & 0.6643 & 0.6877\\ \midrule\midrule
\end{tabular}
}
\end{table}

From Table~\ref{tab:182050} we observe that
BatchNorm-minus achieves slightly better results than BatchNorm. 
Recall that for BatchNorm-minus we found its optimal batch size to be~20,
while~30 was optimal for BatchNorm. 
This suggests that for ResNet18, it may be better to use 
BatchNorm-minus, since a smaller batch sizes have been empirically linked to 
improved convergence properties~\cite{batchsize}.
Also, for ResNet18, our BatchNorm-minus
results provide evidence that the shift and scale parameters 
are not as important as the normalization step.

\subsubsection{ResNet50 Results}

As can be observed by comparing Table~\ref{tab:502050}
to Table~\ref{tab:182050}, we obtain much different results with ResNet50 as compared to
ResNet18. Specifically, 
the performance of BatchNorm-minus dropped dramatically, 
while BatchNorm is the best performer in the ResNet50 case. 
This could be due to the bottleneck architecture of ResNet50,
which projects the convolution layers into a lower dimension before re-projecting 
them into their original dimensions. 

\begin{table}[!htb]
\caption{ResNet50 experimental results}\label{tab:502050}
\centering
\adjustbox{scale=0.85}{
\begin{tabular}{c|cc}\midrule\midrule
\multirow{2}{*}{Normalization} & \multicolumn{2}{c}{Validation accuracy}\\
 & Batch size~20 & Batch size~50\\ \midrule
BatchNorm & 0.7469 & 0.7424\\
AffineLayer & 0.6957 & 0.6986\\
BatchNorm-minus & 0.5597 & 0.6540\\
None & 0.6786 & 0.6939\\ \midrule\midrule
\end{tabular}
}
\end{table}

In contrast to BatchNorm-minus,
the performance of AffineLayer increases for ResNet50, as compared to ResNet18. 
The parameters~$\beta$ and~$\gamma$ 
may be adding expressiveness back to the model.
Lacking those additional training parameters, BatchNorm-minus has 
trouble converging, and even under-performs the case where no normalization is used.

We conjecture that the~$1\times 1$ convolution of the ResNet50 bottleneck architecture
is stripping the model of information, while the additional shift and scale
parameters enable the model to recover 
some of that information. This would explain why the performance of BatchNorm-minus
falls below AffineLayer. Whatever the reason, it appears that for ResNet50, 
the affine parameters play a much more significant role than in ResNet18.
Another takeaway is that AffineLayer, which has no normalization step, 
is clearly outperforming BatchNorm-minus, regardless of the batch size. 
These results indicate that
the normalization step is not the primary driver 
of improved performance in this case. 

Note that the hypotheses that the success of BatchNorm
is due to smoothing the loss surface~\cite{BatchNormnoics} 
or decoupling the length and direction of the weight vectors~\cite{decoupling}
both are related to the normalization step. Since BatchNorm-minus performs
poorly on ResNet50, our results indicate that these hypotheses fail to paint a complete picture, 
at least with respect to ResNet50. In the next section, we shall
see that the same comment holds for ResNet101. Since both of these
architectures use bottleneck blocks---whereas ResNet18 and ResNet34 do not---it is 
likely that this is a key factor.

\subsection{ResNet34 and ResNet101 Experiments}\label{sect:34_101}

In the previous section, we observed that BatchNorm-minus performs well
with ResNet18, but poorly with ResNet50. One possible explanation for this
difference is that the bottleneck block of ResNet50---which is 
not present in ResNet18---benefits from BatchNorm.
To test this hypothesis, in this section we consider a series of experiments
involving ResNet34 and ResNet101.

ResNet34 has the same general structure as ResNet50 
and the same number of residual blocks 
but uses basic blocks instead of bottleneck blocks. 
Thus, other than the difference in depth, the primary
difference between ResNet34 and ResNet50 is the block type.
ResNet101 is a deeper version of ResNet50 with the same bottleneck block structure.

\subsubsection{ResNet34 Results}

Comparing the results for ResNet34 in Table~\ref{tab:342050} 
to those for ResNet18 in Table~\ref{tab:182050},
we see more similarities than differences. 
In both cases, BatchNorm-minus marginally outperforms BatchNorm, 
with the other two cases trailing far behind
This provides additional evidence that the bottleneck block 
affects the way that BatchNorm and its variants work. 

\begin{table}[!htb]
\caption{ResNet34 experimental results}\label{tab:342050}
\centering
\adjustbox{scale=0.85}{
\begin{tabular}{c|cc}\midrule\midrule
\multirow{2}{*}{Normalization} & \multicolumn{2}{c}{Validation accuracy}\\
 & Batch size~20 & Batch size~50\\ \midrule
BatchNorm & 0.7717 & 0.7554\\
AffineLayer & 0.6856 & 0.6837\\
BatchNorm-minus & 0.7719 & 0.7557\\
None & 0.6661 & 0.6782\\ \midrule\midrule
\end{tabular}
}
\end{table}

\subsubsection{ResNet101 Results}

The results in Table~\ref{tab:1012050} for ResNet101 are analogous to what we observed for ResNet50.
In this case, the performance for each normalization is worse than 
ResNet18 or ResNet34 and, crucially, the drop-off for 
BatchNorm-minus is large. AffineLayer is the best performer
in this case. Since AffineLayer includes the two parameters that bring ICS back into the model, this is
additional evidence that ICS reduction is not the reason for the success of BatchNorm. In fact, 
BatchNorm-minus would provide the tightest ICS control, and it gives us the worst results
in this case.

\begin{table}[!htb]
\caption{ResNet101 experimental results}\label{tab:1012050}
\centering
\adjustbox{scale=0.85}{
\begin{tabular}{c|cc}\midrule\midrule
\multirow{2}{*}{Normalization} & \multicolumn{2}{c}{Validation accuracy}\\
 & Batch size~20 & Batch size~50\\ \midrule
BatchNorm & 0.6971 & 0.6746\\
AffineLayer & 0.7032 & 0.6959\\
BatchNorm-minus & 0.4412 & 0.4128\\
None & 0.6819 & 0.6845\\ \midrule\midrule
\end{tabular}
}
\end{table}

\subsection{Analysis of Weights and Gradients}

For ResNet18, we found that BatchNorm clearly outperformed AffineLayer;
see Table~\ref{tab:182050}, above. 
In an effort to better understand the reasons for the differing performance
of these two normalization schemes, 
we extract the weights and gradients for the first epoch of ResNet18. 
Specifically, we consider the first and last layers of the ResNet18 model,
based on the first~20 updates and the last~20 updates.  
We refer to these as ``input'' (first layer), ``final'' (last layer), 
``early'' (first~20 updates), and ``late'' (last~20 updates). 
Here, we extract the weights and gradients for each of 
input-early, input-late, final-early, and final-late, 
and we plot various histograms of these distributions.

\begin{figure}[!htb]
    \centering
    \begin{tabular}{ccc}
        \includegraphics[width=0.4\textwidth]{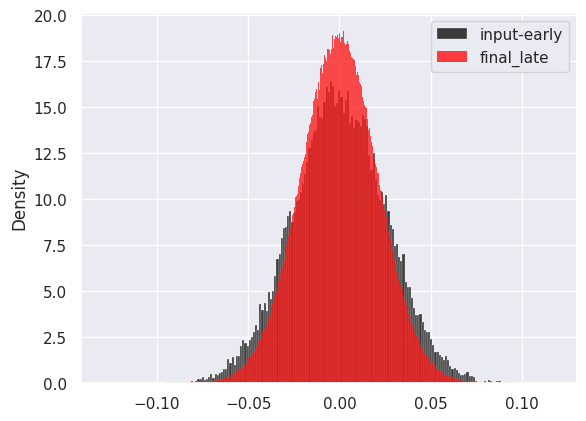}
        & &
        \includegraphics[width=0.4\textwidth]{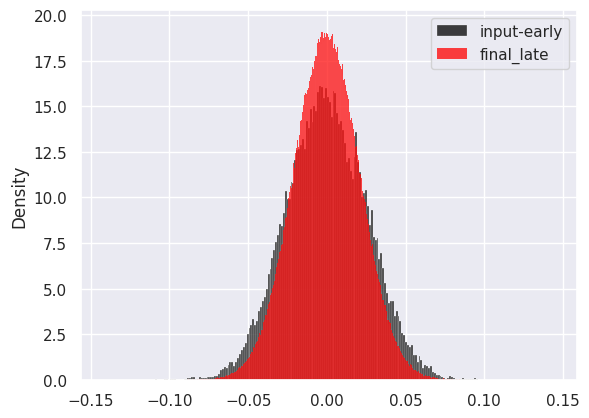}
        \\
        \adjustbox{scale=0.85}{(a) BatchNorm weights}
        & & 
        \adjustbox{scale=0.85}{(b) AffineLayer weights}
        \\ \\[-2.0ex]
        \includegraphics[width=0.4\textwidth]{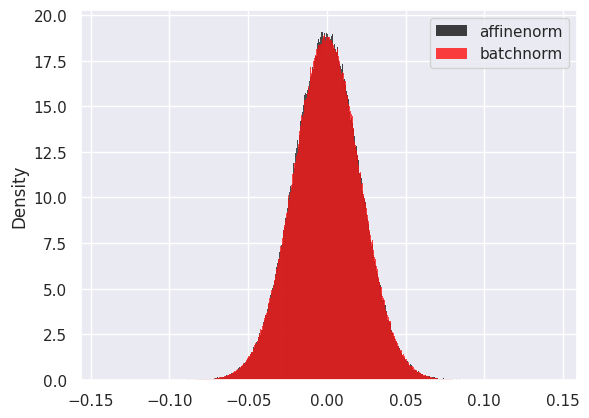}
        & &
         \includegraphics[width=0.4\textwidth]{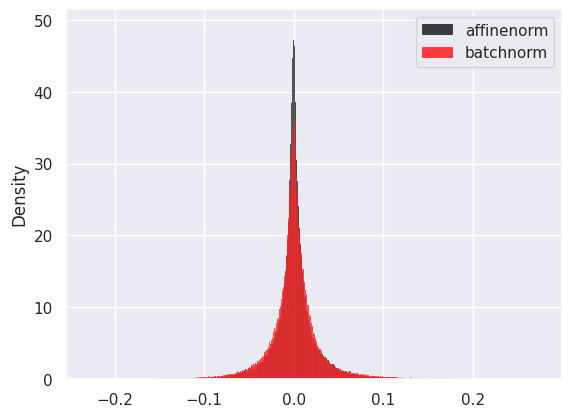}
        \\
        \adjustbox{scale=0.85}{(c) Final-late weights}
        & & 
        \adjustbox{scale=0.85}{(d) Final-late gradients}
    \end{tabular}
    \caption{Weight and gradient comparisons}\label{fig:WG}
\end{figure}

In Figure~\ref{fig:WG}(a),
we see that the weight distribution changes 
from input-early to final-late are fairly modest for BatchNorm,
while Figure~\ref{fig:WG}(b) yields the same conclusion for AffineLayer. 
In Figure~\ref{fig:WG}(c), we have overlayed the final-late weights of BatchNorm and
AffineLayer, while Figure~\ref{fig:WG}(d) gives the analogous
result for the gradients. 
We see that the weights and gradients behave similarly, and hence
whatever is causing BatchNorm to outperform AffineLayer
does not appear to be distinguishable through this weight and gradient
distribution comparison. The bottom line here is
that AffineLayer has a similar effect on the gradients as BatchNorm,
at least in the all-important first epoch. Since AffineLayer performs
significantly worse than BatchNorm, these results cast doubt on
the claim that the success of BatchNorm is due to it stabilizing 
the gradients~\cite{BatchNormics}.

\begin{figure}[!htb]
    \centering
    \begin{tabular}{ccc}
        \includegraphics[width=0.4\textwidth]{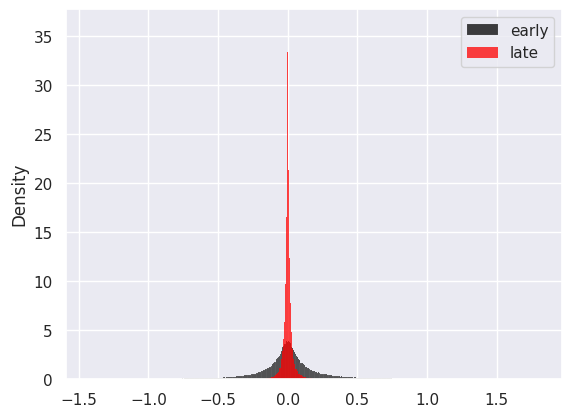}
        & &
        \includegraphics[width=0.4\textwidth]{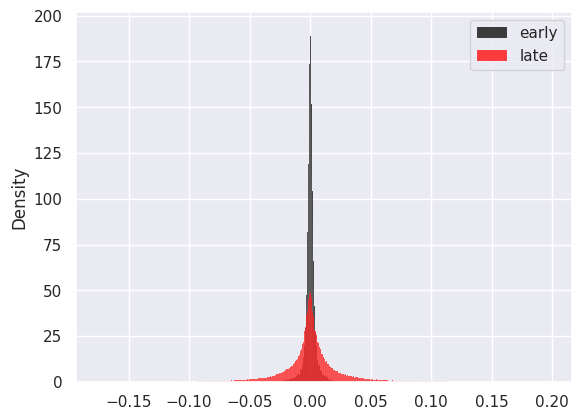}
        \\
        \adjustbox{scale=0.85}{(a) BatchNorm}
        & & 
        \adjustbox{scale=0.85}{(b) AffineLayer}
    \end{tabular}
    \caption{Gradients for input-early vs input-late}\label{fig:Gvs}
\end{figure}

The results in Figure~\ref{fig:WG} show minimal differences
between BatchNorm and AffineLayer. 
A significant difference between AffineLayer and BatchNorm 
can be observed by comparing the input-early and input-late updates.
Specifically, in Figure~\ref{fig:Gvs}(a), we compare the input-early gradient
with the input-late gradient for BatchNorm, while Figure~\ref{fig:Gvs}(b)
provides the analogous comparison for AffineLayer.
We observe from Figure~\ref{fig:Gvs}(a) that for BatchNorm,
the gradient starts spread out and become more tightly focused, 
whereas Figure~\ref{fig:Gvs}(b) shows the opposite behavior
for the AffineLayer gradient. 

Figure~\ref{fig:Gvs} provides additional evidence that it is the weight normalization 
at the beginning of each epoch that is causing the observed performance differences between 
BatchNorm and AffineLayer, at least in the case of ResNet18. We discuss this 
further in the next section.

\subsection{Discussion}

\begin{figure}[!htb]
    \centering
\begin{tikzpicture}[scale=0.75, every node/.style={scale=1.0}]
    \begin{axis}[
        width  = 0.9*\textwidth,
        height = 8cm,
        ymin=0.0,
        ymax=0.8,
        ytick={0.0,0.1,0.2,0.3,0.4,0.5,0.6,0.7,0.8},
        major x tick style = transparent,
        ybar=5*\pgflinewidth,
        bar width=15pt,
        xlabel = {Model},
        ylabel = {Validation Accuracy},
        symbolic x coords={ResNet18, ResNet34, ResNet50, ResNet101},
	y tick label style={
		font=\small,
    		/pgf/number format/.cd,
   		fixed,
   		fixed zerofill,
		1000 sep={},
    		precision=2},
        xtick = data,
        x tick label style={
        		rotate=60,
		font=\small,
		anchor=north east,
		},
        enlarge x limits=0.175,
        legend cell align=left,
        legend pos=south west,
    ]
\addplot[fill=blue,opacity=1.00] 
coordinates {
(ResNet18, 0.7665)
(ResNet34, 0.7717)
(ResNet50, 0.7469)
(ResNet101, 0.6971)
};
\addplot[fill=red,opacity=1.00] 
coordinates {
(ResNet18, 0.6904)
(ResNet34, 0.6856)
(ResNet50, 0.6986)
(ResNet101, 0.7032)
};
\addplot[fill=green,opacity=1.00] 
coordinates {
(ResNet18, 0.7730)
(ResNet34, 0.7719)
(ResNet50, 0.6540)
(ResNet101, 0.4412)
};
\addplot[fill=brown,opacity=1.00] 
coordinates {
(ResNet18, 0.6877)
(ResNet34, 0.6782)
(ResNet50, 0.6939)
(ResNet101, 0.6845)
};
\legend{BatchNorm, AffineLayer, BatchNorm-minus, None}
\end{axis}
\end{tikzpicture}
    \caption{Summary of Results}\label{fig:resultsBN}
\end{figure}
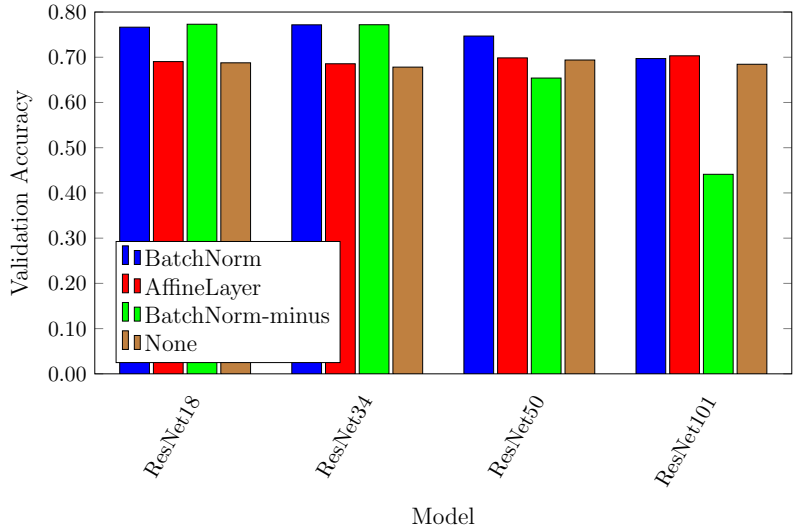

In Figure~\ref{fig:resultsBN}, we summarize the results of our ResNet image classification
experiments. Since we find minimal differences between the results for
batch sizes of~20 and~50, for each model and normalization scheme, we have
graphed the better of the results for batch size~20 or~50.

From Figure~\ref{fig:resultsBN}, we note that for the models tested that employ basic blocks, 
namely, ResNet18 and ResNet34, BatchNorm-minus performs as well as standard BatchNorm.
Since BatchNorm-minus normalizes the weights similarly to BatchNorm,
this is consistent with the results in Figure~\ref{fig:Gvs}, above. Also, 
since BatchNorm-minus lacks the trainable shift and scale
parameters of BatchNorm, it appears that the additional degrees of freedom
provided by these parameters are not particularly useful when training ResNet18 or ResNet34,
and we conjecture that the same is true of any ResNet model that uses basic blocks.
If this is the case, a simpler and more efficient normalization scheme can be used
with such models, without any appreciable loss in performance.

On the other hand, for models tested that include bottleneck blocks, namely, ResNet50 and ResNet101,
BatchNorm-minus performs relatively poorly. Hence, we conclude that the additional
degrees of freedom provided by the shift and scale parameters are critical
for these models, and we conjecture that the same is true of any ResNet
that utilizes bottleneck blocks. Additional evidence that such is the case 
is provided by the fact that AffineLayer---which includes trainable shift and scale 
parameters---performs better on the bottleneck block architectures 
as compared to the basic block architectures.

\section{Conclusion}\label{chap:conclusion}

A considerable body of previous research has focused on BatchNorm, but 
to the best of our knowledge,
none has followed the approach in this paper, where
the normalization and re-parameterization steps are separated and analyzed. 
We applied four distinct normalization schemes to each of ResNet18, ResNet34, ResNet50, 
and ResNet101, and presented a brief analysis of weights and gradients for ResNet18. 

Our main results involve the relative contribution of the 
normalization and re-parameterization steps for the various ResNet architectures tested.
Specifically, we found that for ResNet50 and ResNet101, the trainable shift and scale
parameters appear to increase the expressiveness of the model, allowing it to recover 
more information after the dimensionality reduction step that occurs inside
bottleneck residual blocks. In contrast, for ResNet18 and ResNet34 which use
basic residual blocks, we found that normalization was beneficial, but that 
the additional degrees of freedom provided by shift and scale parameters
did not improve the accuracy. 

We believe that these results for BatchNorm are new and novel,
and provide additional insight into the technique. 
From a practical perspective, our results indicate
that BatchNorm should be used with ResNet architectures that employ 
bottleneck blocks. However, we also found that a simpler and
slightly more efficient technique, BatchNorm-minus, can perform
as well as BatchNorm on ResNet architectures that
use basic residual blocks. When appropriate, the use 
of the simpler BatchNorm-minus normalization could allow for 
smaller batch sizes without sacrificing the speed of convergence.

For future work, it would be interesting to develop more customized optimizers for the two types 
of residual blocks, namely, basic blocks and bottleneck blocks. Given that default implementations of 
ResNet come with BatchNorm built in, it is possible that there is some reduction in performance 
caused by assuming that it is optimal for all types of residual blocks. 
It would also be interesting to consider classifiers where BatchNorm is commonly used
that do not rely on residual blocks, 
and perform similar experiments as presented in this paper. 

\bibliographystyle{plain}

\bibliography{references.bib}

\end{document}